\documentclass[sigconf]{aamas} 

\usepackage{balance} 
\usepackage{algorithm}
\usepackage{algorithmic}
\usepackage{multirow}
\usepackage{subcaption}

\setcopyright{none}
\acmConference[ALA '22]{Proc.\@ of the Adaptive and Learning Agents Workshop (ALA 2022)}
{May 9-10, 2022}{Online, \url{https://ala2022.github.io/}}{Cruz, Hayes, da Silva, Santos (eds.)}
\copyrightyear{2022}
\acmYear{2022}
\acmDOI{}
\acmPrice{}
\acmISBN{}
\acmSubmissionID{???}

\title [Learning to Communicate Using Counterfactual Reasoning]{Learning to Communicate Using Counterfactual Reasoning}

\author{Simon Vanneste}
\affiliation{
  \institution{University of Antwerp - imec \\
               IDLab - Faculty of Applied Engineering}
  \city{Antwerp}
  \country{Belgium}}
\email{simon.vanneste@uantwerpen.be}

\author{Astrid Vanneste}
\affiliation{
  \institution{University of Antwerp - imec \\
               IDLab - Faculty of Applied Engineering}
  \city{Antwerp}
  \country{Belgium}}
\email{astrid.vanneste@uantwerpen.be}

\author{Kevin Mets}
\affiliation{
  \institution{University of Antwerp - imec\\
              IDLab - Department of Computer Science}
  \city{Antwerp}
  \country{Belgium}}
\email{kevin.mets@uantwerpen.be}

\author{Tom De Schepper}
\affiliation{
  \institution{University of Antwerp - imec\\
              IDLab - Department of Computer Science}
  \city{Antwerp}
  \country{Belgium}}
\email{tom.deschepper@uantwerpen.be}

\author{Ali Anwar}
\affiliation{
  \institution{University of Antwerp - imec \\
               IDLab - Faculty of Applied Engineering}
  \city{Antwerp}
  \country{Belgium}}
\email{ali.anwar@uantwerpen.be}

\author{Siegfried Mercelis}
\affiliation{
  \institution{University of Antwerp - imec \\
               IDLab - Faculty of Applied Engineering}
  \city{Antwerp}
  \country{Belgium}}
\email{siegfried.mercelis@uantwerpen.be}

\author{Steven Latré}
\affiliation{
  \institution{University of Antwerp - imec\\
              IDLab - Department of Computer Science}
  \city{Antwerp}
  \country{Belgium}}
\email{steven.latre@uantwerpen.be}

\author{Peter Hellinckx}
\affiliation{
  \institution{University of Antwerp - imec \\
               IDLab - Faculty of Applied Engineering}
  \city{Antwerp}
  \country{Belgium}}
\email{peter.hellinckx@uantwerpen.be}

\begin{abstract}
Learning to communicate in order to share state information is an active problem in the area of multi-agent reinforcement learning (MARL).
The credit assignment problem, the non-stationarity of the communication environment and 
the problem of encouraging the agents to be influenced by incoming messages are major challenges within this research field which need to be overcome in order to learn a valid communication protocol.
This paper introduces the novel multi-agent counterfactual communication learning (MACC) method which adapts counterfactual reasoning in order to overcome the credit assignment problem for communicating agents.
Next, the non-stationarity of the communication environment, while learning the communication Q-function, is overcome by creating the communication Q-function using the action policy of the other agents and the Q-function of the action environment.
As the exact method to create the communication Q-function can be computationally intensive for a large number of agents, two approximation methods are proposed.
Additionally, a social loss function is introduced in order to create influenceable agents, which is required to learn a valid communication protocol.
Our experiments show that MACC is able to outperform the state-of-the-art baselines in four different scenarios in the Particle environment.
Finally, we demonstrate the scalability of MACC in a matrix environment.
\end{abstract}


\begin{document}


\pagestyle{fancy}
\fancyhead{}


\maketitle 


\section{Introduction}
A lot of research has been done towards single-agent reinforcement learning (RL) \citep{mnih2013dqn, mnih2016asynchronous, sutton1998introduction, hasselt2015deep, wang2015dueling}. However, many of the practical applications can naturally be described as cooperative multi-agent systems such as industrial robotics and network packet routing. Single agent RL approaches suffer in this kind of applications as the observation space increases and the joint action space will increase exponentially with the number of agents. 
Cooperative Multi-agent reinforcement learning (MARL) \citep{busoniu2008comprehensive} models these applications as a number of decentralized policies with their individual observation, actions and policy. These decentralized policies are trained using a shared team reward in order to promote cooperation between the agents. While this separation into different agents will help with the scalability issue, the agent can encounter partial observability due to the individual agents only having access to their individual observations. This can make it hard or even impossible to find an effective action policy.
Cooperative multi-agent systems can be extended with inter-agent communication to overcome this issue. The communication allows the agent to share information about their individual observation with other agents. As it is unknown what information is needed by the other agents, the communication between the agents needs to be learned. The learned communication protocol gives the agent access to the parts of the global state which it requires to learn an effective action policy.

In this work, we use the centralized training and decentralized execution (CTDE) paradigm \cite{foerster2016learning, jorge2016learning, foerster2018counterfactual}. This allows us to train the policies in a centralized manner and use additional information (e.g. other policies, observations). After the training of the policies, these policies can be executed decentralized without the need for additional information that is used during training. 
We focus on a non-differentiable binary communication setting (e.g. \{[0, 0], [0, 1], [1, 0], [1, 1]\}) in which it takes one timestep to receive a message which was sent by another agent. This one timestep delayed binary communication is very close to the communication that is used within most real world applications (e.g. digital communication).
However, several problems arise within this setting. First, the agents within a cooperative multi-agent reinforcement learning setting that use a shared team reward will encounter the credit assignment problem \cite{chang2004all, foerster2018counterfactual}. The credit assignment problem occurs when agents are having difficulty deducing what impact an action had on the shared team reward. This problem becomes even larger when learning inter-agent communication as the shared team reward is not only used to learn an action policy, but also to learn a communication protocol.
Next, the non-stationarity problem \cite{foerster2017stabilising} will occur when multiple agent simultaneously learn in a multi-agent setting. This makes the environment appear non-stationary as actions that were previously rewarded at a given state may not be rewarded anymore as the behaviour of the other agents has changed. This problem also increases when learning inter-agent communication because the change in the action policy of the agents changes the utility of a message.

In this paper, we present multi-agent counterfactual communication (MACC) learning, a RL method to simultaneously learn to act and communicate in a multi-agent environment. Multi-agent counterfactual reasoning for the action policy, in order to overcome the credit assignment problem, has already been described and used by \citet{foerster2018counterfactual} in their COMA method. 
However, MACC extends COMA by using counterfactual reasoning to learn a communication protocol with a centralized critic. This critic requires an additional novel communication Q-function which is created from the action Q-function and the policies of the other agents using a novel decomposition into two separate communication Q-functions. This minimizes the non-stationarity of the communication environment for the communication Q-function. As the computational cost of the exact method to calculate the communication Q-function rises exponentially with the number of agents, we investigate two approximation methods. Next, we investigate the impact of communication between two communication policies on the communication Q-function and propose an algorithm to include this.
Additionally, a novel social loss function is introduced for the agent policies in order to promote social behaviour and thereby improve the learning stability. MACC is evaluated and compared against MADDPG \cite{lowe2017multi} and COMA \cite{foerster2018counterfactual} in four different scenarios within the Particle environment from OpenAI \citep{lowe2017multi, mordatch2017emergence}. We chose these methods because they most closely match our approach and the problem setting that we target. Finally, we investigate the scalability of the different MACC variations on a Matrix environment.

This paper is structured as follows. In Section \ref{sec:related_work} of this paper, we discuss relevant literature to our work. Section \ref{sec:background} provides background in Markov Decision Processes and counterfactual reasoning \cite{foerster2018counterfactual}. In Section \ref{sec:methods}, we explain the different components of MACC in detail. Section \ref{sec:experiments} shows the different experiments we performed to demonstrate our method. Finally, our conclusions and future work are described in Section \ref{sec:conclusion}.

\section{Related Work}
\label{sec:related_work}

Recently, several different models for multi-agent communication learning have been presented. The foundations in this research field were laid by \citet{foerster2016learning} and \citet{sukhbaatar2016learning}. \citet{foerster2016learning} proposed two different methods. In Reinforced Inter Agent Learning (RIAL), the communication policy is learned by applying the reward in the same way as we do for the action policy. However, the best results are achieved with Differentiable Inter Agent Learning (DIAL) which uses gradient feedback from the receiving agent to learn the communication policy. \citet{sukhbaatar2016learning} use a similar technique in CommNet. However, CommNet uses continuous communication instead of discrete communication and assumes that multiple steps of communication can occur before the agents take an action. 
These fundamental works were followed by more research in the field of multi-agent communication learning. \citet{peng2017bicnet} presented BiCNet, an actor-critic model that is able to play real-time strategy games such as StarCraft. \citet{mao2017accnet} proposed two methods, one where communication between actors is learned and one where communication between critics is learned. In contrast to our method, both \citet{peng2017bicnet} and \citet{mao2017accnet} use a separate local critic for each of the agents.

A lot of recent work investigates non-broadcast communication. \citet{jiang2018attentional} propose ATOC which allows agents to choose whether communication is necessary for cooperation and which agents to communicate with. TarMAC \citep{das2018tarmac} allows both the sender and the receiver to determine the importance of a certain message. I2C \cite{ding2020learning} is a method which realizes non-broadcast communication by learning a prior network in order to create a belief of the other agents which is used to decide whether to communicate with the other agent or not.
Other recent advances include the work of \citet{ossenkopf2019hierarchical}. They use 
deep MARL to improve the long-term coordination of agents. \citet{vanneste2020} target the lazy agent problem in communication learning by applying value decomposition \citep{sunehag2018value} on DIAL \citep{foerster2016learning}, resulting in improved learning speed and performance.
In the multi-agent deep deterministic policy gradient (MADDPG) method, \citet{lowe2017multi} introduced the idea of using a centralized critic for each agent in MARL . They adapted deep deterministic policy gradient \citep{silver2014deterministic} by using this centralized critic.
\citet{simoes2020} introduce Asynchronous Advantage Actor Centralized-Critic with Communication (A3C3), a method based on the single agent Asynchronous Advantage Actor-Critic (A3C) method \citep{mnih2016asynchronous}. A3C3 uses a centralized critic to make a value estimation of a centralized observation. The communication network is optimized by propagating the gradients of the receiving actors through the communication network. 

\citet{jaques2019social} propose a method that uses social influence as an extra component of the communication reward. The communication policy is trained by a reward composed of the sum of the team reward and a social influence reward. This reward is calculated by determining how much the message influenced the action choice of the other agent. The ideas behind the work of \citet{jaques2019social} and our work seem similar. 
However, there are major differences. \citet{jaques2019social} still use the team reward and purely looks at the change in action distribution. This change in the action distribution does not necessarily result in an improved communication protocol. Instead of using the change in action distribution, we use the action distribution in combination with the action Q-function to determine if a message will result in improved performance.

\section{Background}
\label{sec:background}
In this section, the Markov Decision Process for multi-agent communication and counterfactual reasoning are discussed.
In this work, we use the notation where the superscript indicates the index of a certain agent.
The index uses the notation $a$ for the current agent and $-a$ for every agent except the current agent. The subscript is used to indicate if a symbol is used for the action ($u$) or communication ($c$) environment or to indicate the timestep $t$. The notation $s_{t,t+1}$ is used as an abbreviated notation for $s_{t}, s_{t+1}$. Next, $\mathbb{E}[X;P]$ will represent the expected value of X under the distribution P.

\subsection{Markov Decision Processes}
\begin{figure}[t]
    \centering
    \includegraphics[trim=15 0 0 0,clip, width=0.75\linewidth]{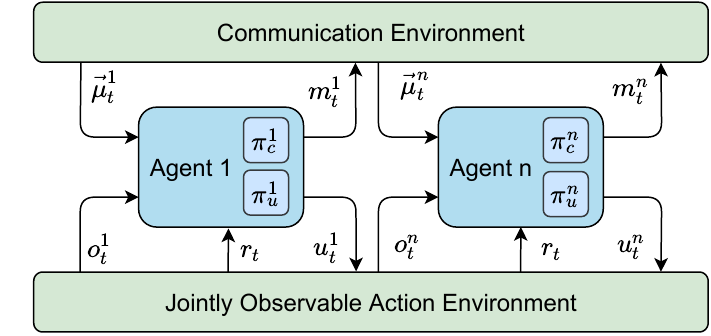}
    \caption{Dec-MDP with a separate action environment and communication environment.}
    \label{fig:dec-MDP}
    \Description{Figure 1. Fully described in the text.}
\end{figure}

In this paper, we use the decentralized Markov Decision Process framework (Dec-MDP) \cite{oliehoek2016concise} extended with communication as shown in Figure \ref{fig:dec-MDP}.
In Dec-MDPs, $n$ agents learn their policy based on the global team reward $r_t$. Every agent only receives a partial observation $o_t$ of the full state $s_t$.
In addition to the jointly observable action environment, a communication environment is added. Each time-step $t$, agent $a$ receives, alongside the observation $o^a_t$ from the action environment, a number of incoming messages $\vec{\mu}^a_t$ from the communication environment. The action policy of the agent $\pi^a_u(u^a_t|o^a_t, \vec{\mu}^a_t)$ takes these inputs and samples an action $u^a_t$. These actions are then processed by the environment and a team reward $r_t$ is given to the agents. The communication policy of the agent $\pi^a_c(m^a_{t}|o^a_t, \vec{\mu}^a_t)$ uses the observation and received messages to generate an outgoing message $m^a_t$. The messages are processed by the communication environment to get the input messages for the next time-step $\vec{\mu}_{t+1} = M(m_t)$. The communication function $M$ determines who receives which messages at the next time-step and $\vec{\mu}^a_{t+1} = M^a(m_t)$ is the part of the communication environment which determines what messages are received by agent $a$. When combining the policies of all agents we can describe the joint action policy $\pi_{u}(u_t | o_t, \vec{\mu}_t)$ and the joint communication policy $\pi_{c}(m_t | o_t, \vec{\mu}_t)$. In our work, we allow for agents that can either send messages, perform actions in the environment or both depending on the requirements of the application.

\subsection{Counterfactual Multi-Agent Policy Gradient}
\citet{foerster2018counterfactual} showed that policy gradient agents in a multi-agent system can be trained using a centralised critic. This centralised critic is able to predict the joint state-action utility $Q_u(s_t, u_t)$ which can be used to calculate the action advantage function $A_u^a$. This calculation uses counterfactual reasoning by subtracting the expected utility of the action policy $V_u^a(s_t, u^{-a}_t)$ from the state-action utility. The expected utility can be expanded into the marginalization over the counterfactual actions $u'^a_t$ of the action policy $\pi^a_u$ which is multiplied with the joint action Q-value of that action permutation. Equation \ref{eq:action_advantage} and \ref{eq:action_advantage_v} show the advantage calculation for the action policy of agent $a$ as described by \citet{foerster2018counterfactual}.
\begin{equation}
A_u^a(s_t, u_t) =  Q_u(s_t, u_t) - V_u^a(s_t, u^{-a}_t)
\label{eq:action_advantage}
\end{equation}
\begin{equation}
V_u^a(s_t, u^{-a}_t) = \sum_{u'^a_t} \Big(Q_u(s_t, (u'^a_t, u^{-a}_t)) \pi^a_u(u'^a_t|o^a_t) \Big)
\label{eq:action_advantage_v}
\end{equation}
The actions of the other agents $u^{-a}_t$ are constant during the marginalization so that the agent only reasons about its own actions. The action advantage function $A^a_u$ calculates the advantage for the action $u^a_t$ using the joint action Q-function. The joint action Q-function is learned by the centralized critic during training.

\section{Methods}
\label{sec:methods}
\begin{figure}[t]
    \centering
    \includegraphics[trim=365 3007 1620 155,clip, width=0.7\linewidth]{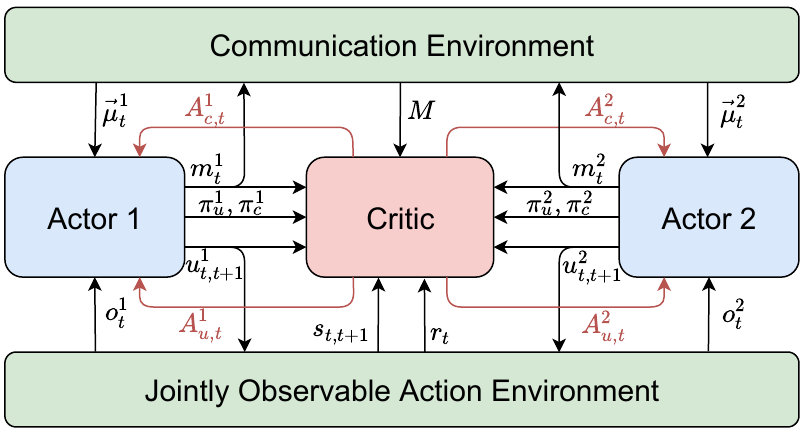}
    \caption{The MACC architecture using a centralised critic.}
    \label{fig:architecture}
    \Description{Figure 2. Fully described in the text.}
\end{figure}

In this section, we discuss the multi-agent counterfactual communication (MACC) learning method. 
Figure \ref{fig:architecture} shows the high level architecture of MACC. 
In MACC, multiple actors act in the environment and communicate with each other. The centralized critic learns a joint action Q-function based on the current state and the received rewards. The critic uses the action Q-function in order to calculate the action and communication advantage for both policies using the action and communication policies of the different agents. 

First, in Section \ref{sec:method_counterfactual_action} we discuss the modifications we made to perform counterfactual reasoning for the action policy to include received messages. Next, the training of the joint action Q-function is described in Section \ref{sec:method_action_q_function}. In Section \ref{sec:method_counterfactual_communication}, we discuss how counterfactual reasoning can be used to calculate an advantage for the communication policies. This advantage calculations requires a communication Q-function which is discussed in Section \ref{sec:qcu}. Section \ref{sec:method_approx} describes approximation methods to calculate the communication Q-function. In Section \ref{sec:qcc}, we discuss the final component of the communication Q-function when including counterfactual reasoning with communication policies that can receive messages. Finally, the novel social loss function is discussed in Section \ref{sec:method_social_loss}.

\subsection{Counterfactual Reasoning in the Action Environment}\label{sec:method_counterfactual_action}
The action policy advantage and value calculations are adapted to include received messages $\vec{\mu}^a_t$ as show in Equation \ref{eq:macc_action_advantage} and Equation \ref{eq:macc_action_advantage_v}.

\begin{equation}
A_u^a(s_t, \vec{\mu}^a_t, u_t) =  Q_u(s_t, u_t) - V_u^a(s_t, \vec{\mu}^a_t, u^{-a}_t)
\label{eq:macc_action_advantage}
\end{equation}
\begin{equation}
V_u^a(s_t, \vec{\mu}^a_t, u^{-a}_t) = \sum_{u'^a_t} \Big(Q_u(s_t, (u'^a_t, u^{-a}_t))\pi^a_u(u'^a_t|o^a_t, \vec{\mu}^a_t)\Big)
\label{eq:macc_action_advantage_v}
\end{equation}
\subsection{Action Q-function}\label{sec:method_action_q_function}
In order to calculate the action policy advantage, the critic needs to learn the joint action Q-function $Q_u(s, u, \theta_u)$ which is represented by the neural network parameters $\theta_u$. This neural network is trained by minimising the loss function from Equation \ref{eq:action_critic_loss}.
\begin{equation}
\begin{aligned}
\mathcal{L}(\theta_{u}^i) = \mathbb{E}_{s_t,u_t,r_t,s_{t+1}}[(r_t + \gamma Q_u(s_{t+1}, u_{t+1}, \theta_{u}^{i-}) - Q_u(s_t, u_t, \theta_{u}^i))^2] 
\end{aligned}
\label{eq:action_critic_loss}
\end{equation}
The loss function uses a target network $\theta_{u}^{i-}$ \cite{mnih2015human}, which is a delayed version of the $\theta_{u}^i$ parameters, in order to stabilize the training of the action Q-function. The loss function uses the observed future actions $u_{t+1}$ (the SARSA update rule as described by \citet{rummery1994line}, \citet{sutton1998introduction}) instead of using the actions that maximize the Q-function ($\max_{u'_{t+1}} Q_u(s_{t+1}, u'_{t+1}, \theta_u)$) in order to minimize the number of inference calls to the action Q-network when the joint action space is large (due to the large number of action permutation over the different agents).
In this work, we use a replay buffer to reuse past experiences to train the joint action Q-function. However, the joint action loss function is on-policy which cannot be used in combination with a replay buffer unless the actions $\widetilde{u}_{t+1}$ are resampled based on the current policy $\widetilde{u}_{t+1} \sim \pi_u(o_{t+1}, M(\widetilde{m}_t))$. Additionally, the MACC action policies can depend on the received messages which depend on the communication policy. This means that the message $\widetilde{m}_t$ also needs to be resampled from the communication policies $\widetilde{m}_t \sim \pi_c(o_t, \mu_t)$.

\subsection{Counterfactual Reasoning in the Communication Environment}\label{sec:method_counterfactual_communication}
We adapted the counterfactual reasoning process to work within the communication environment.
This is shown in Equation \ref{eq:message_advantage} and \ref{eq:message_advantage_v}.
First, the state $s_t$ is expanded to $s_{t, t+1}$ to include $t+1$ which is required when messages take one timestep to arrive. The $t+1$ timestep can be replaced with $t$ for instant communication and with $t+t_d$ for communication delayed by $t_d$ timesteps.
Next, the action Q-function $Q_u(s_t, u_t)$ is replaced by a communication Q-function $Q_c(s_{t, t+1}, m_{t})$.
\begin{equation}
A_c^a(s_{t, t+1}, \vec{\mu}^a_t, m_t) = Q_c(s_{t, t+1}, m_{t}) - V_c^a(s_{t, t+1}, \vec{\mu}^a_t, m^{-a}_t)
\label{eq:message_advantage}
\end{equation}
\begin{equation}
V_c^a(s_{t, t+1}, \vec{\mu}^a_t, m^{-a}_t) = \sum_{m'^a_{t}} \Big(Q_c(s_{t, t+1}, (m'^a_t, m^{-a}_t)) \pi^a_c(m'^a_{t} | o^a_t,\vec{\mu}^a_t)\Big)
\label{eq:message_advantage_v}
\end{equation}
However, the communication Q-function $Q_c$ cannot be learned as the communication environment appears non-stationary because the utility of a message changes when the action policy of the other agent changes during the training process.
In this work, we created the communication Q-function $Q_c$ analytically using the action policies of the other agents and the centralized action Q-function.
The communication Q-function $Q_c$ is composed out of the communication policy to the action policy Q-function $Q_{cu}$ (the impact of a message on other action policies) and the discounted communication policy to the communication policy Q-function $Q_{cc}$ (the impact of a message on other communication policies). This is shown in Equation \ref{eq:qc}. The individual calculations for $Q_{cu}$ are discussed in Section \ref{sec:qcu} and the calculations for $Q_{cc}$ are discussed in Section \ref{sec:qcc}.
\begin{equation}
\begin{aligned}
Q_c(s_{t, t+1}, m_t) = Q_{cu}(s_{t, t+1}, m_t) + \gamma_c Q_{cc}(s_{t, t+1}, m_t)
\end{aligned}
\label{eq:qc}
\end{equation}
The communication discount factor $\gamma_c$ is used to limit the impact of a message on the communication Q-function. The discount factor $\gamma_c$ will be equal to the action discount factor $\gamma$ when using a communication delay of a single timestep. In the general setting the discount factor can be defined using the number of timesteps the communication is delayed $t_d$ and the action discount factor $\gamma$ as $\gamma_c = \gamma^{t_d}$.

\subsection{Communication Policy to Action Policy Q-function}\label{sec:qcu}
The communication policy to action policy Q-function $Q_{cu}$ represents the expected return when a message is sent to the action policy of the different agents. This expectation is shown in Equation \ref{eq:qcu}.
\begin{equation}
Q_{cu}(s_{t, t+1}, m_t) =\mathbb{E}[Q_{u}(s_{t, t+1}, u'_{t+1}); \pi_{u}(u'_{t+1}|o_{t+1}, M(m_t))]
\label{eq:qcu}
\end{equation}
We can uses this expectation to define an exact function for the communication policy to action policy Q-function $Q_{cu}$ as shown in Equation \ref{eq:qcu_exact} which is used in MACC Exact. Intuitively, we will iterate over all possible action permutations $u'_{t+1}$ and multiply the utility for a joint state action pair $Q_{u}(s_{t, t+1}, u'_{t+1})$ with the probability of the joint action being selected $\pi_{u}(u'_{t+1}|o_{t+1}, M(m_t))$ given the input messages $m_t$. The state and actions used in this communication are from $t+1$ as we used one step delayed communication.
\begin{equation}
Q_{cu}(s_{t, t+1}, m_t) = \sum_{u'_{t+1}} \big(Q_u(s_{t+1}, u'_{t+1}) \pi_{u}(u'_{t+1} | o_{t+1}, M(m_t))\big)
\label{eq:qcu_exact}
\end{equation}
However, the computation cost of the exact MACC method rises exponentially with the number of agents as the exact method needs to iterate over all possible action permutations of the other agents. In the next section, we discuss two methods to calculate the communication policy to action policy Q-function $Q_{cu}$ using approximations to reduce this computational cost.

\subsection{Approximation Methods}\label{sec:method_approx}
In this section, the two approximation methods for the communication policy to action policy Q-function $Q_{cu}$ are discussed. These methods aim to reduce the computational cost of counterfactual reasoning for the communication policy for a higher number of agents.
\subsubsection{Sample Mean method}
The Sample Mean method allows us to sample the communication to action policy Q-function under the distribution $\pi_u$ using $n$ samples (see Equation \ref{eq:sampling_mean}). When using a sufficiently high number of samples, the sample mean method will approximate the communication policy to action policy Q-function $Q_{cu} \approx \hat{Q}_{cu}$.
\begin{equation}
\begin{aligned}
&\hat{Q}_{cu}(s_{t, t+1}, m_t) = \dfrac{1}{n} \sum\limits_{i=0}^n Q_u(s_{t+1},\widetilde{u}_{t+1}) \\
& \text{with: } \widetilde{u}_{t+1} \sim \pi_{u}(o_{t+1}, M(m_t)) \\
\end{aligned}
\label{eq:sampling_mean}
\end{equation}
The variance of this estimation will depend on the variance of $Q_u$ under $\pi_u$ and the number of samples $n$ as shown in Equation \ref{eq:sampling_mean_var}. The exact number of samples needs to be determined empirically.

\begin{equation}
VAR[\hat{Q}_{cu}]=\dfrac{VAR[Q_u; \pi_u]}{n}
\label{eq:sampling_mean_var}
\end{equation}

\subsubsection{Agent Based Sampling Method}\label{sec:abs}
The Agent Based Sampling method (ABS) uses the decomposition of the Q-function expectation with respect to the joint action policy $\pi_u$. This decomposition will allow us to compute the two expectations using different methods to find a balance between the exact method and the sampling mean method.
\begin{lemma}
The communication policy to action policy Q-function $Q_{cu}$ expectation under $\pi_u$ can be decomposed into a Q-function expectation with respect for $\pi^{a'}_u$ and $\pi^{-a'}_u$ for any agent $a'$.
\begin{equation}
\begin{aligned}
Q_{cu}&(s_{t, t+1}, m_t) = \\
\mathbb{E}[\mathbb{E}[&Q_{u}(s_{t, t+1}, (u'^{a'}_{t+1}, u'^{-a'}_{t+1})); \pi_{u}^{a'}(u'^{a'}_{t+1} | o^{a'}_{t+1}, M^{a'}(m_t))];\\
&\pi_{u}^{-a'}(u'^{-a'}_{t+1} | o^{-a'}_{t+1}, M^{-a'}(m_t))]
\end{aligned}
\label{eq:agent_based_sampling}
\end{equation}
\end{lemma}
\begin{proof}
\begin{equation}
\begin{aligned}
Q_{cu}&(s_{t, t+1}, m_t) \\
=&\mathbb{E}[Q_{u}(s_{t, t+1}, u'_{t+1}); \pi_{u}(u'_{t+1}|o_{t+1}, M(m_t)] \\
=&\sum_{u'_{t+1}} \big(Q_u(s_{t+1}, u'_{t+1}) \pi_{u}(u'_{t+1} | o_{t+1}, M(m_t))\big) \\
=&\sum_{u'^{-a'}_{t+1}} \sum_{u'^{a'}_{t+1}} \big(Q_u(s_{t+1}, (u'^{a'}_{t+1}, u'^{-a'}_{t+1}))\cdot \\
&\pi_{u}^{a'}(u'^{a'}_{t+1} | o^{a'}_{t+1}, M^{a'}(m_t)) \pi_{u}^{-a'}(u'^{-a'}_{t+1} | o^{-a'}_{t+1}, M^{-a'}(m_t))\big) \\
=&\sum_{u'^{-a'}_{t+1}} \big( \mathbb{E}[Q_{u}(s_{t, t+1}, (u'^{a'}_{t+1}, u'^{-a'}_{t+1})); \pi_{u}^{a'}(u'^{a'}_{t+1} | o^{a'}_{t+1}, M^{a'}(m_t))] \\
&\pi_{u}^{-a'}(u'^{-a'}_{t+1} | o^{-a'}_{t+1}, M^{-a'}(m_t))\big) \\
=&\mathbb{E}[\mathbb{E}[Q_{u}(s_{t, t+1}, (u'^{a'}_{t+1}, u'^{-a'}_{t+1})); \pi_{u}^{a'}(u'^{a'}_{t+1} | o^{a'}_{t+1}, M^{a'}(m_t))];\\
&\pi_{u}^{-a'}(u'^{-a'}_{t+1} | o^{-a'}_{t+1}, M^{-a'}(m_t))]
\end{aligned}
\label{eq:agent_based_sampling_proof}
\end{equation}
\end{proof}
The sampling mean will be used for the expectation over the other agents $-a'$ and the expectation over agent $a'$ will be calculated using the exact method as show in Equation \ref{eq:agent_based_sampling_with_importance_sampling}. When using a sufficiently large number of samples $n$ the agent based sampling estimation will approximate communication policy to action policy Q-function $Q_{cu} \approx \hat{Q}_{cu}$. Note that we select a different agent for every sample in order to prevent a certain bias towards any agent.
Intuitively, this is similar to how humans will reason about the reaction of other humans about a certain event. They will reason about how every individual will react instead of going over all the possible permutations. Note that this method will also be used to efficiently calculate the communication policy to communication policy Q-function $Q_{cc}$ as discussed in the next section.

\begin{equation}
\begin{aligned}
\hat{Q}_{cu}(s_{t, t+1}, m_t) = \dfrac{1}{n} \sum\limits_{a'=0}^n \mathbb{E}[Q&_{u}(s_{t, t+1}, (u'^{a'}_{t+1}, u'^{-a'}_{t+1})); \\
\pi&_{u}^{a'}(u'^{a'}_{t+1} | o^{a'}_{t+1}, M^{a'}(m_t))]\\
= \dfrac{1}{n} \sum\limits_{a'=0}^n \sum&_{u'^{a'}_{t+1}} Q_u(s_{t+1},(u'^{a'}_{t+1}, \widetilde{u}_{t+1}^{\:-a'})) \cdot \\
& \pi_{u}^{a'}(u^{a'}_{t+1} | o^{a'}_{t+1}, M^{a'}(m_t))\\
\text{with: } \widetilde{u}_{t+1}^{\:-a'}& \sim \pi_{u}^{-a'}(o^{-a'}_{t+1}, M^{-a'}(m_t))\\
\end{aligned}
\label{eq:agent_based_sampling_with_importance_sampling}
\end{equation}
The variance of the ABS estimation is shown in Equation \ref{eq:agent_based_sampling_with_importance_sampling_var}. Note that this variance will be lower as the variance will be under $\pi_u^{-a'}$ instead of under $\pi_u$. However, a sample from the ABS method is more computationally expensive. Nevertheless, we hypothesise that the variance will be reduced because we can directly use the distribution $\pi_u^{a'}$ in our estimation instead of sampling this distribution.
\begin{equation}
VAR[\hat{Q}_{cu}]=\dfrac{ VAR[Q_{cu}; \pi_u^{-a'}]}{n}
\label{eq:agent_based_sampling_with_importance_sampling_var}
\end{equation}
In our experiments the number of samples $n$ will be equal to the amount of agents. Hence, the computational cost grows linearly instead of exponentially when using the exact method.

\subsection{Communication Policy to Communication Policy Q-function}\label{sec:qcc}
The communication policy to communication policy Q-function $Q_{cc}$ is defined as shown in Equation \ref{eq:qcc}.
\begin{equation}
Q_{cc}(s_{t, t+1}, m_t) =\mathbb{E}[Q_c(s_{t+1,t+2},m'_{t+1}); \pi_c(m'_{t+1}|o_{t+1}, M(m_t))] \\
\label{eq:qcc}
\end{equation}
However this poses a computational challenge as in order to compute $Q_{cc}$ at timestep $t$, we need to calculate $Q_c$ at timestep $t+1$ for which we again need to calculate $Q_{cc}$ at timestep $t+1$. This can be overcome by starting the calculations for a certain episode (with an episode length of $T$) at timestep $t=T-2$ (as $Q_{cc}=0$ at timestep $t=T-1$ because of the one step communication delay) and calculate $Q_{cc}$ backwards towards $t=0$. Algorithm \ref{alg:algorithm} shows the full MACC algorithm including the $Q_{cc}$ calculations.
\begin{algorithm}[tb]
\caption{MACC}
\label{alg:algorithm}
\begin{algorithmic}
\STATE Initialise $\theta_u, \theta_u^-, \theta_{\pi_u}, \theta_{\pi_c}$
\FOR{each episode e}
    \STATE $s_0 =$ initial state, t = 1
    \STATE $\mu_0^a = 0$ for each agent $a$
    \WHILE{$s_t \ne$ terminal \AND $t < T$}
        \FOR{each agent $a$}
            \STATE $u_t^a \sim \pi_u^a(o_t^a, \mu_t^a, \theta_{\pi_u^a})$, $m_t^a \sim \pi_c^a(o_t^a, \mu_t^a, \theta_{\pi_c^a})$
        \ENDFOR
        \STATE $\vec{\mu}_{t+1} = M(m_t)$
        \STATE Get $s_{t+1}, r_t$ from the environment
        \STATE $t=t+1$
    \ENDWHILE
    \STATE Calculate $Q_u^a, Q_{cu}^a$ for each agent $a$
    \STATE $Q_{c, t-1}^a = 0$ for each agent $a$
    \FOR{j = t - 2 \TO 0}
        \FOR{each agent $a$}
            \STATE Calculate $Q_{cc, j}^a$ using $Q_{cu, j+1}^a, Q_{c, j+1}^a$
            \STATE $Q_{c, j} = Q_{cu, j} + \gamma_c Q_{cc, j}$
        \ENDFOR
    \ENDFOR
    \STATE Calculate $A_u, A_c, V_u, V_c$ using $Q_u, Q_c$
    \STATE Update $\theta_u, \theta_u^-$ using $r$ and $\theta_{\pi_u}, \theta_{\pi_c}$ using $A_u, A_c$
\ENDFOR
\end{algorithmic}
\end{algorithm}
Note that we cannot use the exact method or the sampling mean method because we do not calculate the $Q_c$ value for every message combination $m'_{t+1}$ but only for the alternative message $m'^{a'}_{t+1}$ from the different agents $a'$. This means that we need to use the agent based sampling method as defined in section \ref{sec:abs} which leads to the Equation \ref{eq:qcc_abs}.
\begin{equation}
\begin{aligned}
Q_{cc}(s_{t, t+1}, m_t) \approx \dfrac{1}{n} \sum\limits_{a'=0}^n \sum_{m'^{a'}_{t+1}} \big(& Q_c(s_{t+1,t+2}, (m'^{a'}_{t+1}, m^{-a'}_{t+1})) \cdot\\
& \pi^{a'}_c(m'^{a'}_{t+1}|o^{a'}_{t+1}, M^{a'}(m_t)) \big)
\end{aligned}
\label{eq:qcc_abs}
\end{equation}

\subsection{Social Loss}\label{sec:method_social_loss}
The MACC methods use the change in the future expected reward through the choice of alternative actions in function of an input message to learn a communication policy. However, when the action policy ignores the received messages, it becomes impossible to learn a valid communication protocol which in turn does not give the action policy any incentive to use the received messages. In order to promote social behaviour from the action policy, which in this context means taking different actions based on the messages of other agents, an optional social loss function $\mathcal{L}^s$ is added to the action policy loss function. This loss function encourages the action policy to have a different policy distribution for different received messages $\vec{\mu}'$.
The social loss function is shown in Equation \ref{eq:social_loss}. This loss function iterates over the different input bits and compares the output distribution, given the original message with the output distribution, with the given input message with a bit flip $\neg\vec{\mu}_x$ for the selected bit $x$.
The social loss function can be adapted to use different loss functions like the 
Kullback-Liebler loss function. However, we found that the absolute difference performed better due to the linear nature of the loss function.
\begin{equation}
\begin{aligned}
\mathcal{L}^s(\theta_{\pi_u^a}^i) =  - \frac{\lambda}{k} \sum_{x=0}^k |\pi_{u}^{a}(o^{a}, \vec{\mu}, \theta_{\pi_u^a}^i) - \pi_{u}^{a}(o^{a}, (\neg\vec{\mu}_x, \vec{\mu}_{-x}), \theta_{\pi_u^a}^i)|
\end{aligned}
\label{eq:social_loss}
\end{equation}
Note that this social loss is an optional loss function which can be controlled by the hyperparameter $\lambda$ which allows us to tune the social loss loss in order to encourage social behaviour without removing the possibility for the agent to ignore the messages when this is required in a certain environment. Additionally, if the sending agent cannot improve the team reward by sending different messages, the communication policy learns to always send the same message in order to promote the best action distribution.

\section{Experiments}
\label{sec:experiments}
In these experiments, MACC is evaluated with and without the social loss ($\lambda = 0$), with different approximation methods and compared with COMA \cite{foerster2018counterfactual} and the RLlib MADDPG \citep{lowe2017multi} implementation because these methods target the credit assignment problem, use a centralized critic and can be used to train discrete non-differentiable inter-agent communication.
The RLlib framework \cite{liang2018rllib} is used to train the different methods efficiently. We used the same hyperparameters for MADDPG as in the work of \citet{lowe2017multi} and determined the hyperparameters for the other methods empirically and by using a grid search.
The results for these methods are the average of five training runs for every method with the best performing run and the least performing run removed. The methods are evaluated on multiple scenarios for both the particle environment \cite{lowe2017multi} and on the custom matrix environment. These scenarios are selected to validate different multi-agent communication configurations.

\subsection{Particle Environment}
\begin{figure}[t]
    \centering
    \includegraphics[width=0.47\textwidth]{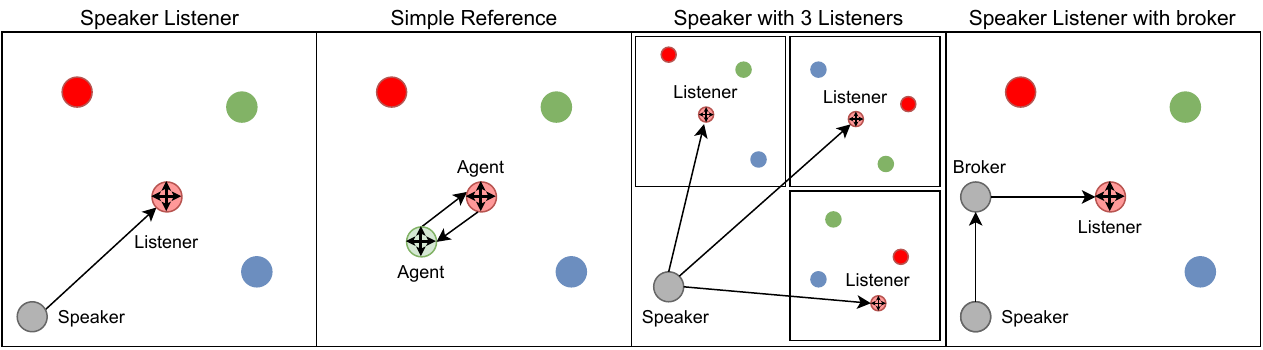}
    \caption{The four different Particle environment scenarios in which the arrows represent the communication topology.}
    \label{fig:particle_env}
    \Description{Figure 3. Fully described in the text.}
\end{figure}

Our experiments use different scenarios from the particle environment \cite{lowe2017multi} (see Figure \ref{fig:particle_env}) where multiple agents are simulated in a 2D environment. We use the Speaker Listener scenario and the Simple Reference scenario from this environment. The goal in these scenarios is that the agent or both agents, for the Simple Reference environment, move towards the goal landmark. But the agents do not know which of the landmarks is the target landmark as this information is only available to the other agent. The agents need to learn to communicate this information and to use this information to move to the target landmark.
Additionally, we propose two more complex novel scenarios, the Speaker with 3 Listeners scenario and the Speaker Listener with Communication Broker scenario. The Speaker with 3 Listeners scenario extends the Speaker Listener scenario with additional listeners that have to navigate to their own landmark. These agents will have the same target landmark color but the landmarks have a different location. The Speaker Listener with Communication Broker scenario is also an extension of the Speaker Listener scenario in which an additional agent will learn to pass messages from the speaker to the listener as the speaker and listener are not allowed to communicate directly increasing the communication complexity.
The agents in these different scenarios use a shared team reward which represents the average distance between the agents and their the target landmark.

\begin{figure*}
\begin{minipage}[b]{.49\textwidth}
    \centering
    \includegraphics[width=0.96\textwidth]{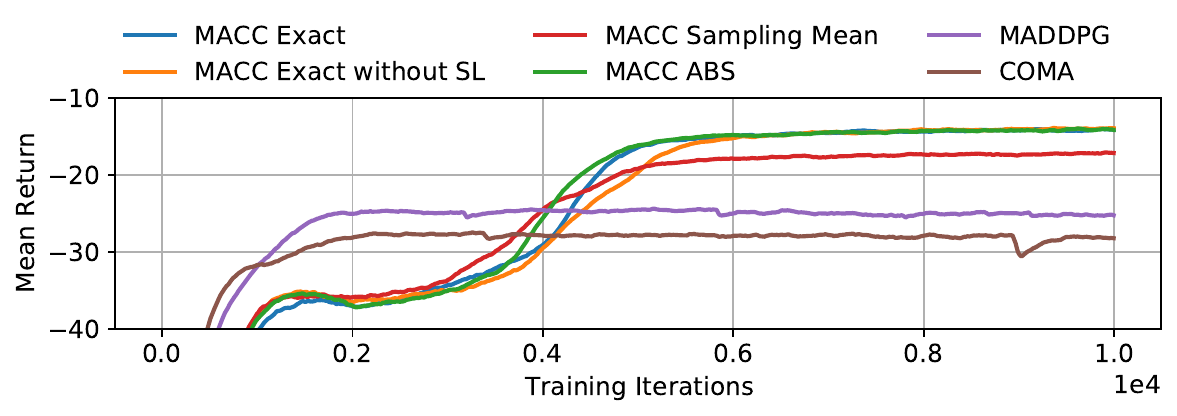}
    \caption{The Speaker Listener training results.}
    \label{fig:speaker_listener}
    \Description[Figure 4. The results from the Speaker Listener experiments.]{The results for the Speaker Listener experiment show that the MACC Exact method is able to learn a valid communication protocol and achieves a reward of -14 after 5000 training iterations. The MACC ABS method performs almost identical as the exact method reaching a reward of -14 after 5000 training iterations. The MACC Sampling Mean method also performs similar to the exact method but gets stuck in a local optimum and can only achieve a reward of -17 after 5000 training iterations. MACC without the social loss function can achieve a reward of -14 but it can only achieve this after 6000 training iterations. The MADDPG method achieves a reward of -25 after only 2000 training iterations. COMA can also reach the highest reward after 2000 training iterations but can only achieve a reward of -28.}
\end{minipage}
\begin{minipage}[b]{.49\textwidth}
    \centering
    \includegraphics[width=0.96\textwidth]{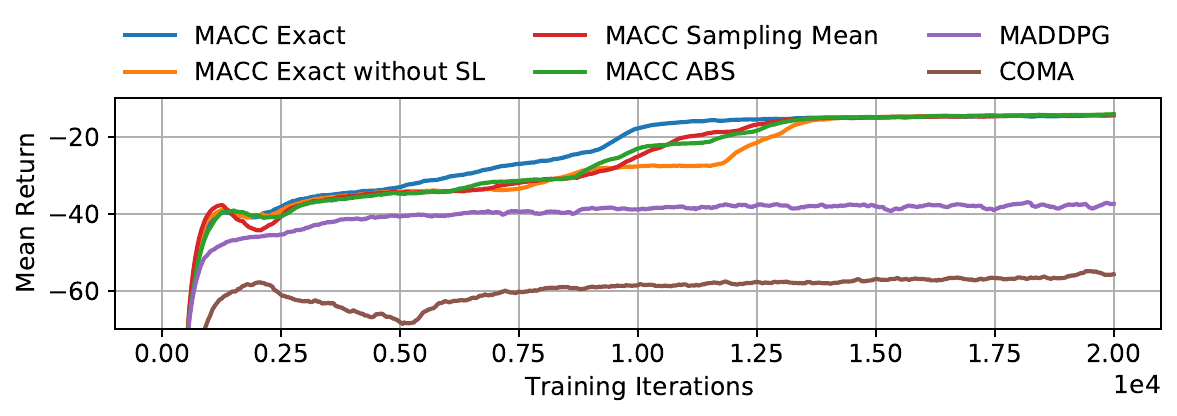}
    \caption{The Simple Reference training results.}
    \label{fig:simple_reference}
    \Description[Figure 5. The results from the Simple Reference experiments.]{The results for the Simple Reference experiment show that the different MACC versions are all able to learn a valid communication protocol and achieve a reward of -14. However, the exact method is able to learn this after 10000 training iterations. Both MACC ABS and Sampling Mean require 13000 training iterations and MACC without the social loss functions needs 14000 training iterations. MADDPG cannot learn a valid communication protocol and reaches a reward of -40 after 5000 training iterations. Finally, the COMA method is only able to obtain a reward of -60.}
\end{minipage}
\end{figure*}
\begin{figure*}
\begin{minipage}[b]{.49\textwidth}
    \centering
    \includegraphics[width=0.96\textwidth]{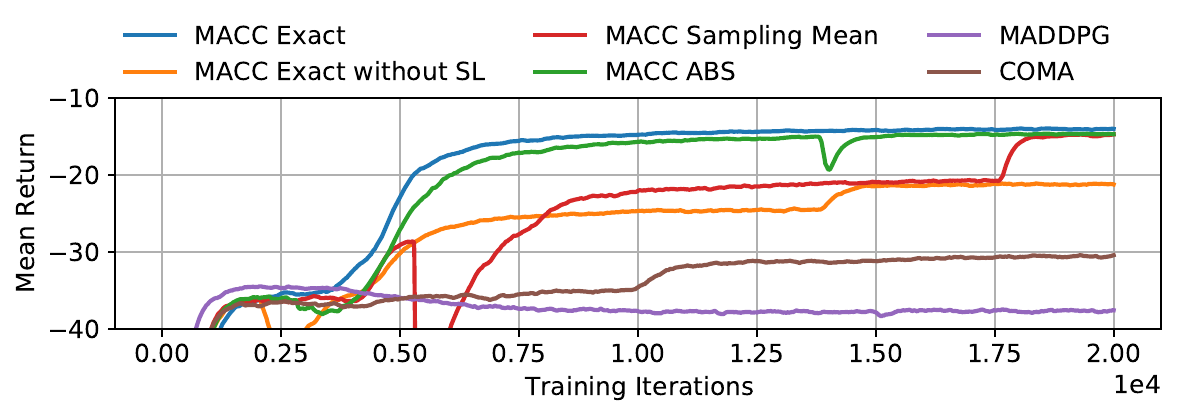}
    \caption{The Speaker with 3 Listeners training results.}
    \label{fig:speaker_3_listener}
    \Description[Figure 6. The results from the Speaker with 3 Listeners experiments.]{The results from the Speaker with 3 Listeners experiment show that MACC Exact and MACC ABS are able to obtain a reward of -14 after 10000 training iterations. The MACC Sampling Mean needs 19000 training iterations to achieve the same result. MACC Exact without the social loss is able to reach an average reward of -21 after 15000 training iterations. The COMA method achieves a reward of -30 after 12500 training iterations. MADDPG reaches a reward of -35 after 2500 training iterations but cannot learn a valid communication protocol.}
\end{minipage}
\begin{minipage}[b]{.49\textwidth}
    \centering
    \includegraphics[width=0.96\textwidth]{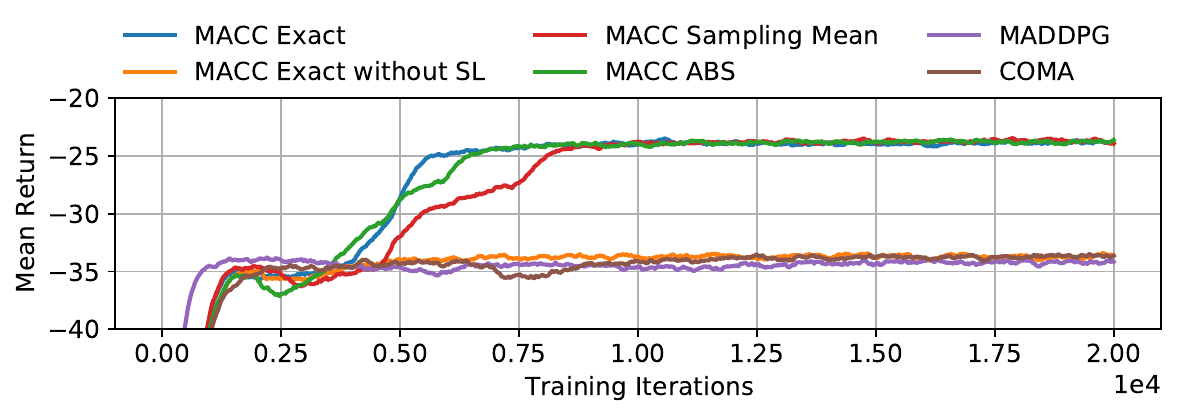}
    \caption{Speaker listener with broker training results.}
    \label{fig:speaker_listener_hub}
    \Description[Figure 7. The results from the Speaker listener with broker experiments.]{The results from the speaker listener with broker experiment show that MACC Exact, MACC ABS and MACC Sampling mean achieve a reward of -23. MACC Exact and MACC ABS achieve this after 6000 training iterations while MACC sampling mean needs 8000 training iterations. MADDPG, COMA and MACC without the social loss function are not able to learn a communication protocol and only reach a reward of -35.}
\end{minipage}
\end{figure*}

\begin{table*}[]
\caption{The reward statistics for the final 10\% of the training run within the Particle environment.}
\label{tab:results}
\begin{tabular}{lllllll}\toprule
Experiment name & MACC Exact & MACC Exact & MACC Sampling & MACC ABS & COMA & MADDPG \\
 & & without SL & Mean & & & \\
\midrule
Speaker Listener & $-14.10 \pm 1.04$ & $-13.95 \pm 1.09$ & $-17.15 \pm 4.55$ & $-14.11 \pm 1.12$ & $-28.17 \pm 8.58$ & $-25.19 \pm 7.20$ \\
Simple Reference & $-14.56 \pm 0.81$ & $-14.39 \pm 0.76$ & $-14.51 \pm 0.81$ & $-14.32 \pm 0.79$ & $-56.03 \pm 2.85$ & $-37.80 \pm 2.47$ \\
Speaker with 3 Listeners & $-14.01 \pm 0.72$ & $-21.18 \pm 3.62$ & $-14.82 \pm 0.72$ & $-14.66 \pm 0.81$ & $-30.51 \pm 4.85$ & $-37.62 \pm 1.25$ \\
Speaker Listener with broker & $-23.94 \pm 1.50$ & $-33.74 \pm 1.76$ & $-23.71 \pm 1.42$ & $-23.74 \pm 1.48$ & $-33.79 \pm 1.71$ & $-34.21 \pm 1.85$ \\
\bottomrule
\end{tabular}
\end{table*}

\subsubsection{Speaker Listener}\label{sec:speaker_listener}
The training results of the different methods for the Speaker Listener scenario are shown in Figure \ref{fig:speaker_listener} and Table \ref{tab:results}. These results show that every MACC configuration is able to outperform both COMA and MADDPG. They are able to learn a basic communication protocol (a higher reward than -35) but cannot achieve a reward higher then -25. Note that COMA and MADPPG learn this basic communication faster than MACC is able to learn a communication protocol. This is due to the way these methods learn their communication protocol. COMA and MADDPG learn their communication protocol directly from the critic while MACC will use the critic in combination with the policies of the other agents to learn a communication protocol. The policies of the other agents need a certain amount of time to learn to change their behaviour based on the input message. However MACC is able to learn a valid communication protocol and achieve a reward of -15. MACC using the exact method and MACC ABS achieve a very similar learning behaviour while MACC Exact without the social loss function takes longer to learn the communication protocol and MACC Sampling Mean achieves a lower overal reward

\begin{figure*}
     \centering
     \begin{subfigure}[b]{0.33\textwidth}
         \centering
         \includegraphics[width=\textwidth]{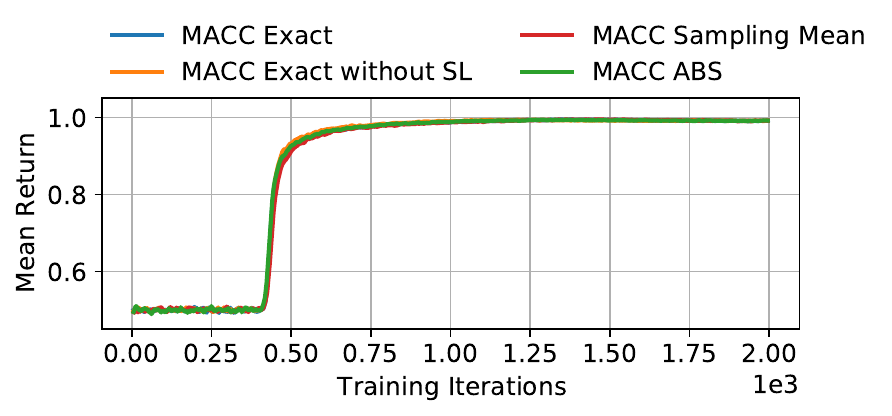}
         \caption{2 agents}
         \Description[Figure 8a. The results from the Matrix environment with 2 agents experiments.]{The results from the Matrix environment with 2 agents experiment show that the different variants of MACC perform very similar. They all achieve a reward of 0.99 after 10000 training iterations.}
    \label{fig:matrix_2agents_2bits_2labels}
     \end{subfigure}
     \hfill
     \begin{subfigure}[b]{0.33\textwidth}
         \centering
        \includegraphics[width=\textwidth]{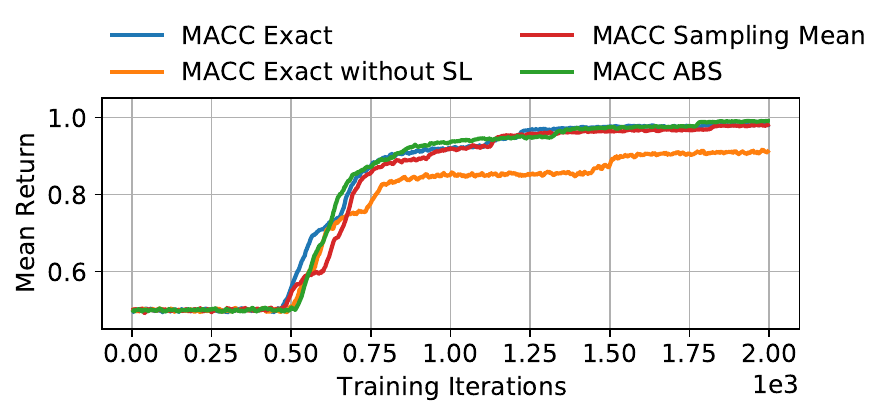}
        \caption{4 agents}
        \label{fig:matrix_4agents_2bits_2labels}
        \Description[Figure 8b. The results from the Matrix environment with 4 agents experiments.]{The results from the Matrix environment with 4 agents experiment show that MACC Exact, MACC ABS and MACC Sampling Mean are all able to achieve 0.98 after 17500 training iterations. MACC Exact without the social loss function can only achieve a reward of 0.91 after 15000 training iterations.}
     \end{subfigure}
     \hfill
     \begin{subfigure}[b]{0.33\textwidth}
         \centering
        \includegraphics[width=\textwidth]{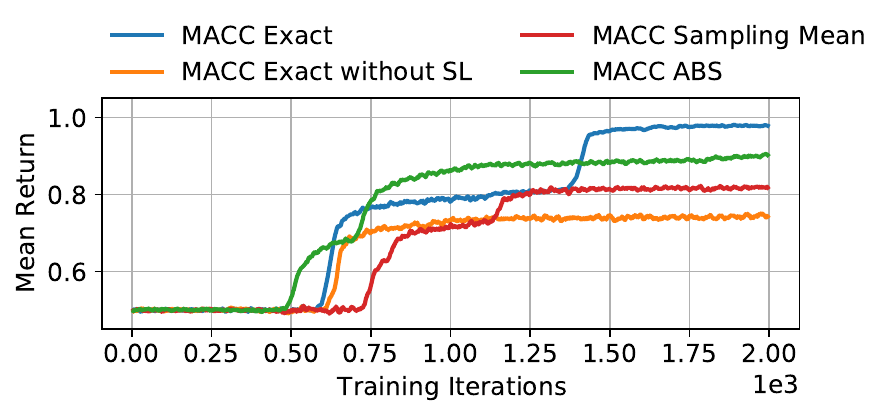}
        \caption{6 agents}
        \label{fig:matrix_6agents_2bits_2labels}
        \Description[Figure 8c. The results from the Matrix environment with 6 agents experiments.]{The results from the Matrix environment with 6 agents experiment show that MACC Exact is able to reach a maximum reward of 0.98 after 15000 training iterations. MACC ABS achieves a reward of 0.9 after 12500 training iterations. The MACC Sampling Mean method reaches a reward of 0.82 after 12500 training iterations. Finally, the MACC Exact method without social loss can only achieve a reward of 0.74 after 10000 training iterations.}
     \end{subfigure}
        \caption{The matrix environment for different number of agents.}
        \label{fig:three graphs}
\end{figure*}

\begin{table*}[]
\caption{The reward statistics for the final 10\% of the training run within the Matrix environment.}
\label{tab:results_matrix}
\begin{tabular}{lllll}\toprule
Experiment name & MACC Exact & MACC Exact without SL & MACC Sampling Mean & MACC ABS \\ \midrule
Matrix with 2 agents & $0.99 \pm 0.00$ & $0.99 \pm 0.00$ & $0.99 \pm 0.00$ & $0.99 \pm 0.00$ \\
Matrix with 4 agents & $0.98 \pm 0.02$ & $0.91 \pm 0.11$ & $0.98 \pm 0.02$ & $0.99 \pm 0.00$ \\
Matrix with 6 agents & $0.98 \pm 0.02$ & $0.74 \pm 0.20$ & $0.82 \pm 0.13$ & $0.90 \pm 0.11$ \\ \bottomrule
\end{tabular}
\end{table*}

\subsubsection{Simple Reference}\label{sec:simple_reference}
The results for the simple reference scenario are presented in Figure \ref{fig:simple_reference} and Table \ref{tab:results}. MADDPG is able to successfully learn a valid action policy but is not able to learn a communication protocol since the average reward is too low. COMA is not able to learn a valid action or communication policy because of learning instability caused by the non-stationary communication environment, the credit assignment problem and the large action space of the agents.
The large actions space is the number of actions times the number of possible output messages ($5 * 2^2 = 20$) as COMA does not have a separate communication policy.
However, since MACC splits the action and communication policy, the different versions of MACC are able to learn a valid communication protocol. These results show that MACC Exact is able to achieve the maximum reward the fastest. The Sampling Mean and ABS approximations are performing similar but require more training iterations to achieve the maximum reward. Finally, MACC Exact without social loss takes the most training iterations to achieve the maximum reward due to the policies taking longer to change their action or message distribution based on the received message.

\subsubsection{Speaker with 3 Listeners}\label{sec:speaker_3_listeners}
The results for the Speaker with 3 Listeners scenario are shown in Figure \ref{fig:speaker_3_listener} and Table \ref{tab:results}. The COMA and MADDPG methods are not able to achieve the same results as in the Speaker Listener scenario. As both the action and communication protocol are learned based on the centralized critic, we see that COMA and MADDPG are not able to scale to learning a protocol with multiple receiving agents. MACC Exact without social loss is not able to achieve the maximum reward as the method reaches a local optimum because the action policies are not encouraged to learn social behaviour. The other MACC variants are able to learn the maximum reward. Note that the MACC Sampling Mean method shows a more unstable learning process compared to MACC Exact and MACC ABS. MACC Sampling Mean will sample actions from all the other agents which can lead to a high variance when the amount of agents increases.

\subsubsection{Speaker Listener with Communication Broker}\label{sec:speaker_broker_listeners}
The results for the speaker listener with communication broker are shown in Figure \ref{fig:speaker_listener_hub} and Table \ref{tab:results}. In this experiment, COMA, MADDPG and MACC Exact without SL are not able to learn a valid communication protocol. The other MACC variants are able to learn a basic communication protocol but can only achieve a maximum reward of -25. The MACC Exact and MACC ABS are performing very similary while MACC Sample Mean takes a lot more training iterations to achieve the same reward.

\subsection{Matrix Environment}
The Matrix environment is based on the work of \citet{lowe2019pitfalls}. Here, every individual agent will receive a distinct natural number within a certain range (0 or 1 in this experiment) with a 50\% chance that all the agents received the same number. The goal of the agent is to determine if they received the same number or not. For every agent that correctly determines if the agents have the same number, the agents receives a team reward of $1/n$ for a configuration with  $n$ agents. So if all the agents are correct, a team reward of 1 is received. This scenario is a many to many communication topology which increases the communication complexity.

The goal of this experiment is to increase the number of agents in order to study the scalability of the different MACC versions. In this experiment, we validate three different configurations for the Matrix environment (2, 4 and 6 agents). As the policies of the different agents are identical to each other, we share the parameters between the different agents.
The results for the different configurations in the Matrix environment are shown in Figure \ref{fig:three graphs} and Table \ref{tab:results_matrix}. These results show the scaling of the different MACC versions with an increasing number of agents. The Exact MACC method achieves the highest reward over the different configurations. The more computationally efficient MACC ABS method achieves similar results as the exact version. The Sampling Mean MACC achieves slightly lower performance although it has the same computational cost as the ABS version. Finally, the MACC Exact without social loss learns a basic communication protocol (the reward is higher than 0.5) but achieves the lowest average reward. This shows the importance of the social loss in a setting with a higher number of agents in order to promote social behaviour.

\section{Conclusion}
\label{sec:conclusion}
In this paper, we present a novel method for multi-agent communication learning called MACC which is able to reason about the impact of a certain action or message using counterfactual reasoning by using a centralised critic. The critic is able to reason about the impact of a certain message on both the action and communication policy. Additionally, we also introduce two approximation methods for MACC and a social loss function in order to improve the social behaviour.
In the experiments, the different variants of MACC are compared to MADDPG and COMA, which also use a centralized critic, in four different scenarios in the Particle environment. These experiments show that the MACC variants are able to learn a valid communication protocol in a range of different communication scenarios and can outperform both MADDPG and COMA. The social loss can be used by MACC to promote social behaviour in the receiving agent. Additionally, we show that the approximation methods are able to learn a valid communication protocol. We observed that MACC Agent Based Sampling is able to slightly outperform MACC Sampling Mean as the number of agents is increased.
Next, the scaling performance of the MACC variants are validated on the Matrix environment. These results show that MACC Exact and MACC Agent Based Sampling outperform the other MACC variants when the number of agents is increased.
In this work, we also assumed to have full access to the policy of every agent. This requirement could be relaxed by learning a model of the policy of the other agents. Additionally, we also believe that the social loss function could be improved in order to further limit the chance of getting stuck in a local optimum, which is still possible in certain environments as demonstrated in the Speaker Listener with broker agent environment.

\begin{acks}
This work was supported by the Research Foundation Flanders (FWO) under Grant Number 1S94120N and Grant Number 1S12121N. We gratefully acknowledge the support of NVIDIA Corporation with the donation of the Titan Xp GPU used for this research.
\end{acks}

\bibliographystyle{ACM-Reference-Format} 
\bibliography{bibliography}

\end{document}